\documentclass{article}

\usepackage{PRIMEarxiv}

\usepackage[utf8]{inputenc} 
\usepackage[T1]{fontenc}    
\usepackage[colorlinks=true, linkcolor=black, citecolor=black, urlcolor=black]{hyperref}
\usepackage{url}            
\usepackage{booktabs}       
\usepackage{amsfonts}       
\usepackage{nicefrac}       
\usepackage{microtype}      
\usepackage{lipsum}
\usepackage{fancyhdr}       
\usepackage{graphicx}       
\graphicspath{{media/}}     

\title{KIRETT: Knowledge-Graph-Based Smart Treatment Assistant for Intelligent Rescue Operations
\thanks{\noindent\textit{Accepted and published in the LWDA'23: Lernen, Wissen, Daten, Analysen. \newline \hspace*{1.5em} Copyright for this paper by its authors.\newline
\hspace*{1.7em}Use permitted under Creative Commons License Attribution 4.0 International (CC BY 4.0).}}
}

\author{
  Mubaris Nadeem, Johannes Zenkert, Lisa Bender, Christian Weber, Madjid Fathi \\
  Institute for Knowledge-Based Systems and Knowledge Management \\
  University of Siegen\\
  Hoelderlinstrasse 3, 57068 Siegen, Germany\\
  \texttt{\{Mubaris Nadeem\}mubaris.nadeem@uni-siegen.de} \\
}

\begin{document}
\maketitle

\begin{abstract}
Over the years, the need for rescue operations throughout the world has increased rapidly. Demographic changes and the resulting risk of injury or health disorders form the basis for emergency calls. In such scenarios, first responders are in a rush to reach the patient in need, provide first aid, and save lives. In these situations, they must be able to provide personalized and optimized healthcare in the shortest possible time and estimate the patient's condition with the help of freshly recorded vital data in an emergency situation. However, in such a time-dependent situation, first responders and medical experts cannot fully grasp their knowledge and need assistance and recommendation for further medical treatments. To achieve this, on the spot calculated, evaluated, and processed knowledge must be made available to improve treatments by first responders. The Knowledge Graph (KG) presented in this article as a central knowledge representation provides first responders with an innovative knowledge management that enables intelligent treatment recommendations with an artificial intelligence-based pre-recognition of the situation.
\end{abstract}

\keywords{Wearable Device \and Artificial Intelligence \and Rescue Operations \and Knowledge Graph \and Treatment Assistance}

\section{Introduction}
Rescue operations are described as an important treatment within the healthcare system, as they are essential to support patients in need of fast and effective treatment in life-threatening situations. In today’s world, however, the amount of rescue operations is increasing rapidly, due to multimorbid health disorders and a higher risk of injuries, with a much higher post-impact on patients’ health. Arriving first responders should provide accurate, fast, and reliable medical assistance in a short time. Their decisions are based on their accessible knowledge, assisted by medical data that indicates health disorders during “treatment”-time. With the accumulated knowledge health professionals can make personalized decisions in a scenario where personal data of patients is crucial. The KIRETT project presents a wearable, which provides an accumulation of treatment knowledge, medical vitals, and situation detection (SD). Having such medical devices at hand, rescue operators can prepare, reconsider, and provide fast and reliable medical treatment. For the treatment knowledge, a knowledge graph (KG) is developed to assist rescue operators in their daily work and support them with contextual recommendations. As the central core of the KIRETT project, the developed KG is fed with information from an artificial neural network (ANN) providing a situation detection (SD), as well as vitals from medically certified devices such as the Zoll X-Series. A simplified user interface (GUI) is a unique communication interface supporting the first responder to allow easy interaction with the embedded device. The recommendation provided by the KG is visualized in text and needs active confirmation from the health professional through clicking on the worn wearable. This enables an additional safety layer for the patient and first responders at hand. This paper focuses on the knowledge management based development of the KG for KIRETT. It mainly focuses on how the KG is defined and integrated with other modules of the project, and will in addition explore how the representation of nodes and relationships was conceptualised for this specific graph. It further presents an evaluation plan, and introduces an explainability component for the project. The following section will first give an overview of the KIRETT project and related publications, and then delve into the specifics of the project.

\subsection{The KIRETT project}
Today's world is characterized by rapid advancements in artificial intelligence (AI) and intelligent systems. This technology and these systems open up new possibilities and methods in various fields. One of these areas is emergency rescue, where innovative technologies can help to save lives. The KIRETT project has set this objective for itself and seeks to combine the derivation KG-based treatment recommendation with AI-based situation recognition. For the KIRETT project, this means that the knowledge of the rescue service in terms of treatment procedures is managed and represented as a graph and is used for the implementation of a demonstrator in the form of a wearable. In a treatment scenario, patients in need are connected to medical-certified devices (e.g. ZOLL X-Series), which provide the medical experts with valuable data that is also fed to the KIRETT device. The wearable is worn by first responders treating patients, and does not need any direct contact with the patients themselves. The scope of the project foresees the possibility of treatment support in mass-causality scenarios (MCI), where the general idea is to connect multiple patients to one device to provide accurate treatment recommendations for each patient. These are depending on the situation, allowing warnings to pop up on the screen if the condition of a patient is worsening. Within this project, various research papers have been published. Table \ref{table:related_work} provides a comprehensive overview of existing publications. 

\begin{table}
  \caption{Related work in the KIRETT project }
  \label{table:related_work}
  \begin{tabular}{ccl}
    \toprule
    Topic & Description & Source\\
    \midrule
        Overview  & General overview of the project &  \cite{overviewZenkert} \\
        Knowledge Graph  & KG approach based on a transferable framework & \cite{abu2022transferrable} \cite{frameworkAbuRasheed} \\
        UX-design & User experience and design of KIRETT wearable & \cite{uxNadeem} \\
        ANN & Application and execution of ML algorithms on FPGA & \cite{cardioShad} \cite{respiratoryShad} \cite{neuroShad} \cite{cardioMike} \\
    \bottomrule
  \end{tabular}
\end{table}

\section{Background}

\subsection{Knowledge Graph} \label{kg}

KGs are defined as knowledge bases that are built upon a graph-like structure to store information on real-world entities and concepts and their relation to one another \cite{Tommasini2020}. Underlying is the concept of knowledge bases, which contents are potentially derived from large scale data sources, which are connected through associated metadata to ensure a sufficient knowledge representation \cite{Tommasini2020}. This can help in use cases ranging from search engines (e.g. DBPedia, Google KG), over recommender systems to question answering systems such as Alexa \cite{Tommasini2020}. In healthcare, KGs are becoming increasingly popular, providing organizations access to medical research and treatment methods to optimize healthcare delivery by helping to validate diagnosis and create treatment plans catering to individual requirements \cite{abu2023healthcare}. A KG consists of nodes representing entities that are connected to each other such as objects, people, places or complex concepts. The connections are the relationships of the graph, called edges, which are creating meaning through specifying the semantic association between two nodes. The graph, depending on the specific domain tailoring and the associated level of formal definition, can be used as a flexible structure for the representation and utilization of different data. As such, it yields a high potential for integration and implementation in different projects, adjusting to the intended usage and structural requirements. Some of such design decisions can be whether to make the graph directional, meaning that relations unidirectional or bidirectional to allow to express bidirectional relations within a graph. A complete graph, as an alternative approach that allows cycles, is a graph where all nodes are connected to one another through combinations of relations. Relationships, therefore, exist between every possible node of the graph, giving a very complete picture of its data and relations from one data point to the next. This may, however, not always be possible to achieve. Besides directions of relationships, cycles and the completeness of a graph, there is also the matter of connectivity. A connected graph does not have subgraphs that are not linked by edges. Starting from one node, one has to be able to reach all other nodes when ignoring the set direction of relationships. A directed graph is considered weakly connected, if a graph is considered only fully connected if the direction of edges is ignored. \cite{borowiecki2006graph}. In the following, the KIRETT graph is categorized in regard to these features, and a mathematical definition will be explored.

\section{Methods} \label{Methodology}
\subsection{The KIRETT Knowledge Graph}
As discussed in \ref{kg} KG, a graph for the project can be built in many different ways. To fully support the functionality and basis that KIRETT is founded on, which is the operation manual for rescue services \cite{treatmentpdf} provided as the data the graph should be constructed from. The KG for the project is a directional, cyclic and weakly connected graph, fully customized to the needs of the project and its integrated data. It can be mathematically described as $KG = \{V,E\}$ where a node (or vertex $V$) can connect through an edge $E$ to another node. The KG used to visualize the treatment paths is a directed graph to implement the order of actions in the medical care, provided through standard operating procedures of rescue services \cite{treatmentpdf}. An undirected graph would not implement any sort of order to treatment procedures, and potentially endanger patient lives by implementing an undesired freedom of choice. The graph is weakly connected, but there are no disconnected subgraphs since graph nodes are only to be accessed by relationships which are keeping the logic behind connections between certain treatments intact. Treatment paths not reachable in the graph would serve no purpose in a procedural order. The graph is furthermore not complete, since not every node is connected to every other node as many nodes do not have common context. 

\section{Implementation and Results}
In the following the implementation of such a KG is presented alongside its interconnections with the embedded modules and platform. The creation of the graph, as well as its chosen implementation regarding node types, relationships and properties will be elaborated on.

\begin{figure}[h]
\centerline{\includegraphics[width=\textwidth]{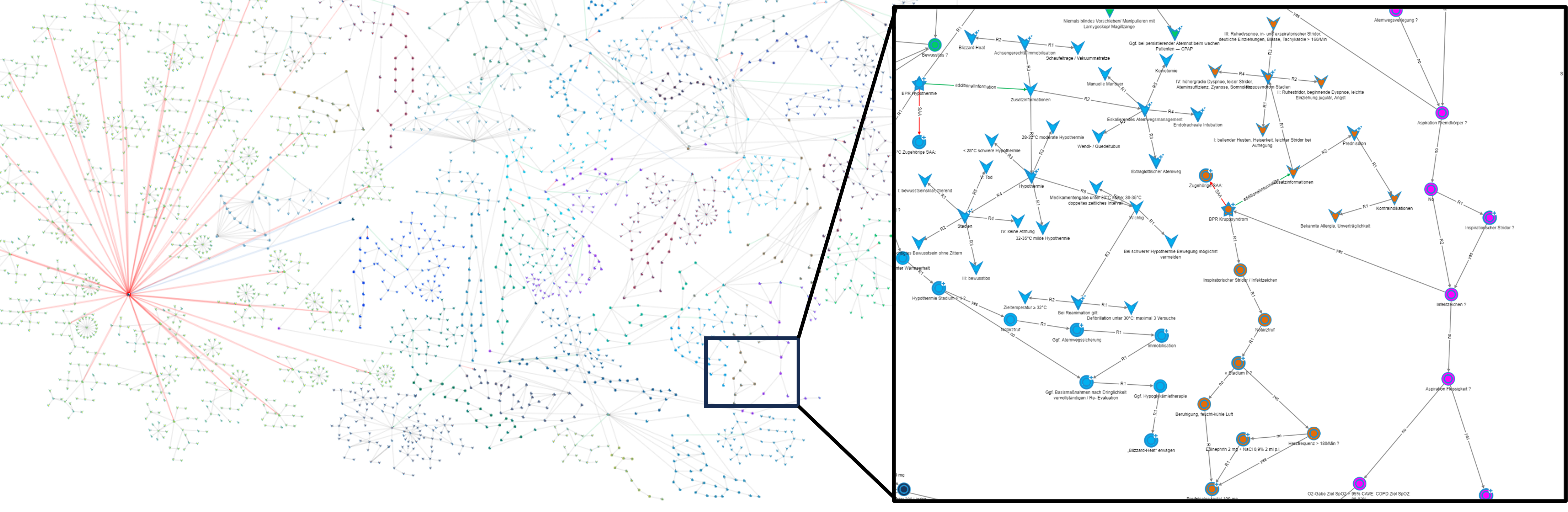}}
\caption{This figure shows the fully expanded KG. On the left side, the whole graph is visible in its complexity, while on the right side of the figure a detailed, zoomed-in version of an arbitrary part of the graph is presented to show the interconnection of treatment possibilities. The various colors of the nodes are used to allow for a better overview. The graph is in German language, due to its data being extracted from a German rescue manual. }
\label{kgfull}
\end{figure}

\subsection{Interconnection of Knowledge Graph and Modules}

The KG (Fig. \ref{kgfull}, left) is defined as the very heart of the KIRETT project, as it holds all the data of the treatment paths (Fig. \ref{kgfull}, right) and standard procedures extracted from the manual for rescue operations \cite{treatmentpdf} used by the rescue teams during treatment. Surrounding it, there are other interconnected modules the KG interacts with, namely the Situation Detection (SD), the Middleware, and the graphical user interface (GUI). All of these modules together make KIRETT what it is. The SD is directly connected to values of a questionnaire, which the KG triggers along the processing of the path options within the KG. After receiving data from the KG, the SD, will evaluate the underlying vital data using an ANN to indicate to the KG the most likely patient situation, or scenario, which informs the treatment path. This will then be processed by the KG, and the rescue personnel will receive a recommendation regarding the group of treatment paths that are most likely to be applicable in the moment. When the KG requires a value such as the patient’s temperature to make a decision or inform the user of the wearable, which is the first responder, the KG will request data through the Middleware to enable the KG to conclude on certain directions of treatment. In such cases the KG sends a request to the Middleware, which is connected to the Zoll X-Series reading patient vitals. These will be retrieved through a database connection and provided to the KG. Using the recorded data, the KG will either make a decision, or relay the values and their meaning for the current treatment to the healthcare personnel. There are many decisions the KG cannot make on its own due to restrictions in sensors, missing information, or the lack of non-formalized human knowledge and human awareness. These will then be relayed to the GUI to inquire the user for a decision or a piece of information. The GUI, alongside the LCD \cite{uxNadeem} it utilizes, is a further module in the KIRETT construct. It is the main source of KG interactions and prompts. The GUI \cite{uxNadeem} is its interaction partner and its window to commune with the healthcare professional using the KIRETT system. Any selected treatment node will pass through the GUI to be shown to the user, and any questions or sudden changes will also be displayed by the GUI. The interconnection between the GUI, the KG and all further modules is done via a messenger system, which allows all components to send messages to each another. Every module has a queue it periodically retrieves and processes messages from. In all of this, the KG is at the center of the model, and has therefore has been constructed in such a way that it enables an optimal workflow for rescue personnel. The following sections will elaborate on the construction and implementation of the graph and its nodes, edges and properties.

\subsection{Building the Graph} \label{graph_building}

\begin{figure}[h]
\centerline{\includegraphics[width=\textwidth]{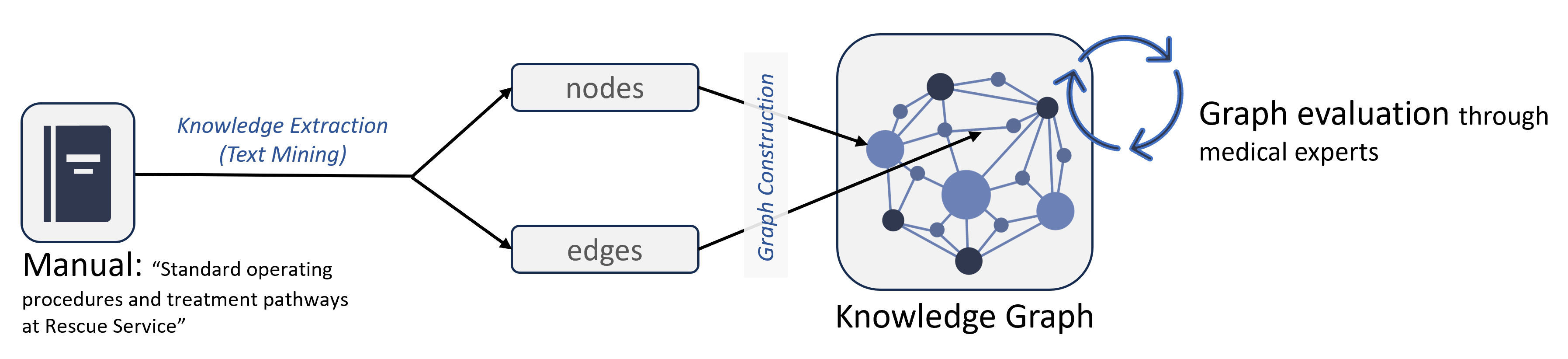}}
\caption{The KG construction pipeline: This figure presents an abstract view of the construction pipeline. First Text Mining was utilized to extract all treatment information, like treatment paths and medication, then manual modeling was implemented due to difficulties with this approach. The knowledge was consequently differentiated between nodes and edges and was constructed into a KG. The subsequently constructed KG was evaluated in regular meetings with the medical experts to ensure a high-quality standard for medical applications.}
\label{pipeline}
\end{figure}

The KG was created based on the manual for rescue operations \cite{treatmentpdf}. This document has been used by the rescue teams during treatments as a standardized catalogue regarding their procedures. The graph has therefore been modeled on its basis, to allow the utmost possible correctness of any containing procedures. At the beginning of the planning phase of the project, the manual was used to determined the required node types based on features of the document, which will be elaborated on further below. After this, knowledge was extracted (Fig. \ref{pipeline}, knowledge extraction). It was manually converted into a node format that was previously agreed on within the implementation team. The information was manually modeled into the KG, and questions about the content and its implementation were noted. Subsequently, the questions were conveyed to the medical partners (Fig. \ref{pipeline}, graph evaluation), and wishes for certain variants of implementation and answers about medical questions were then again used to improve the graph implementation and modeling. In the following, the modeling and implementation details on how nodes, relationships and properties were modeled, will be elaborated.

\begin{figure}[h]
\centerline{\includegraphics[width=\textwidth]{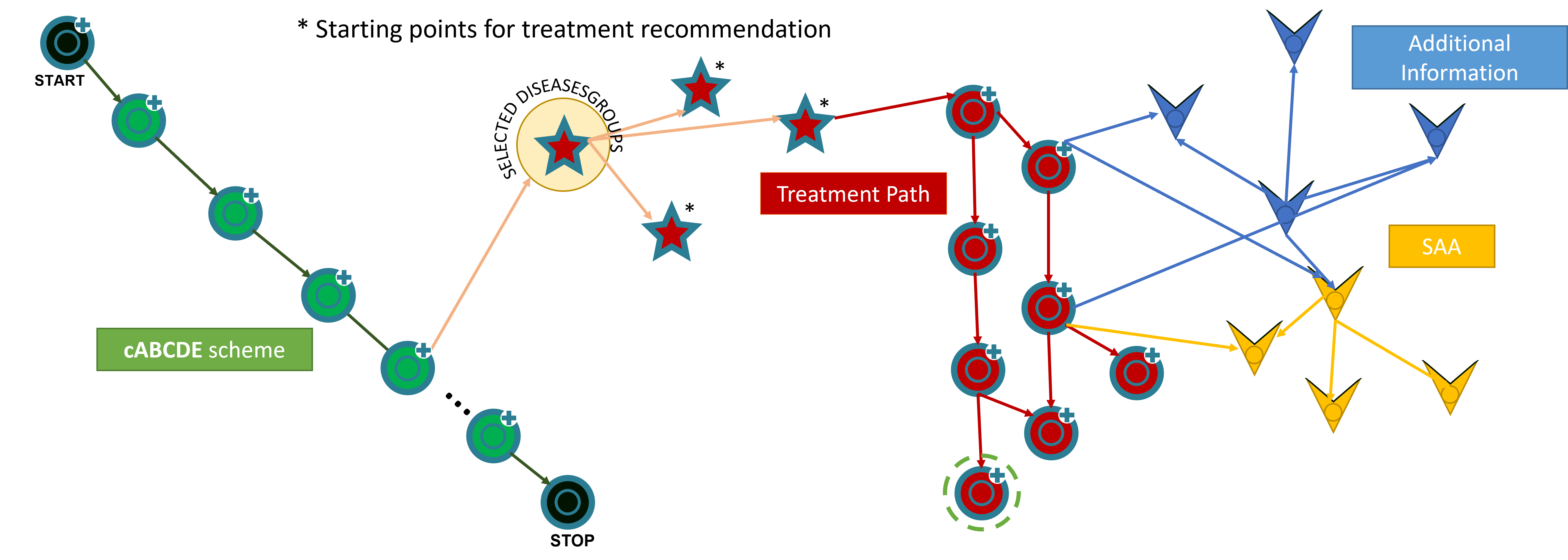}}
\caption{Schematic representation of the KG: This figure shows an abstract representation of the presented KG. Colors are used to allow a different view of different node types and paths. The green node-path is described as the "cABCDE scheme". Red stars present the "selected disease groups". The red circular nodes on the right represents possible treatment paths, connected to the selected disease group. Additional information (blue) and the SAA (yellow) are nodes providing further information on medications, contraindications, doses and warnings.}
\label{scheme}
\end{figure}

\subsection{Nodes} \label{nodetypes}

\begin{table}
  \caption{Node Types}
  \label{table:nodetypes}
  \begin{tabular}{ccl}
    \toprule
    Name & Description \\
    \midrule
    StartNode / StopNode            & Marks the start and end of the treatment\\
    BPRNode / SAANode             & Marks the beginning of a new treatment path or standard procedure\\
    JumpNode             & Node providing shortcuts to different areas of the graph \\
    DecisionNode         & Node indicating a decision is to be made\\
    (Invasive)ProcedureNode    & Node indicating an (invasive) procedure has to take place\\
    ActionNode           & Node indicating an action has to be taken\\
    DisplayNode / WarningNode          & Node displaying potentially critical information\\
    \bottomrule
  \end{tabular}
\end{table}
The KG used in the context of KIRETT, has various node types to navigate the different treatment paths efficiently. Table \ref{table:nodetypes} provides a list of all existing node types with a short description. Depending on the node type, the KG and GUI will interact differently with them. Some nodes may enable easier jumping to other areas of the graph, while some may trigger prompts to the user. The StartNode is the root of the graph, which all treatment starts from. From the StartNode (Fig. \ref{scheme}, green circled node),  the rescue personnel can either follow the cABCDE scheme or follow the questionnaire provided by the SD to help it find an entry point for them. The antithesis to the StartNode is the StopNode, which is where the treatment of the current patient is terminated. This finds most application in cases of multiple patients during a rescue operation, allowing for one treatment to continue, while another one stops. Besides the Start- and StopNode, which indicate the beginning and end of the graph, there are also BPRNodes and SAANodes, which indicate the start of a treatment path (BPR), or a standard procedure path (SAA) (Fig. \ref{scheme}, yellow). They can be utilized to navigate to treatments that are specifically requested by the user or the SD. JumpNodes on the other hand are a way to navigate the KG by jumping from one path to another or connecting certain node types with each other. To allow for optimal connectivity, there are a few special JumpNodes that help navigating the KG by providing shortcuts. There are three primary JumpNodes: BPR-JumpNode, SAA-JumpNode and DiseaseGroup-JumpNode (Fig. \ref{scheme}, red star), which connect to all the respective BPRs, SAAs and DiseaseGroups for better and faster navigation to new treatments, should they be required. Relationships between these primary jump nodes and the respective nodes they connect to (SAANodes, BPRNodes and Disease Groups) are bidirectional to improve interconnectivity between various treatment paths.  Unlike JumpNode, Start/StopNodes and BPR/SAANodes, the following node types were mainly created on bases of the manual for rescue operations \cite{treatmentpdf}, used by the rescue teams during treatments. It contains EPC-like depictions of treatment paths, which were then translated into various node types suiting this graph application based on the differentiation partially already given in the manual. DecisionNodes are the most common nodes to be found throughout the graph, and required a decision to be made by the rescue personnel operating the wearable, often supported by the graph receiving information from the Middleware. DecisionNodeYN indicates a binary decision, the relationships adjacent to it always have the values ”yes” and ”no”. An example for such a node would be: ”Does the patient have fever?”. A DecisionNodeOR, on the other hand, is a node type with more than two attached nodes. The decision here needs to be made on the DecisionNodeOR itself as well as its children. An example for a DecisionNodeOR would be ”Ausculation findings?”, with the possible replies being ”Rhonchus / Buzzing / Spasticity / Expiratory stridor” as well as ”Crackles / coarse crackles” and ”Inconspicuous / Not clear”. The user then has to decide which option fits their findings on the patient best. Besides decisions, procedures also play a great role in medical care. ProcedureNodes indicate the necessity of a medical procedure often referencing a standard procedure of rescue services, while InvasiveProcedureNodes make it apparent that the procedure required is invasive. These nodes have to be executed with the utmost precision and care, and while keeping further information in mind. This additional data is provided directly at the ProcedureNodes through a relationship, as well as an interlink with the standard medical procedure (SAA) that may be required at this point in time to ensure the safety of the patient. ProcedureNodes typically only have one child node, and require no decision making. And example of such a ProcedureNode would be: ”i.v. access”. For other nodes in the graph that do not fit the above categories, an ActionNode is used to indicate the necessity of an action to be taken.  This node also required no decision making, and differentiates from procedures by being less medically-involved, but is still thoroughly important to the treatment itself. An example of such a node would be ”transport to clinic” or “call an additional doctor”. DisplayNodes and WarningNodes are nodes outside of the path that merely serve the purpose of displaying information and warnings at the recue personnel’s volition. They are not required to complete a path, but rather serve as a help to look up information on procedures if necessary.

\subsection{Relationships} 
\begin{table}
  \caption{Relationship Types}
  \label{table:relationshiptypes}
  \begin{tabular}{ccl}
    \toprule
    Relationship Type Name & Description \\
    \midrule
    R\textsuperscript{n}    & Leads to the possible next nodes along the path\\
    BPR / SAA                     & Leads to a related BPRNode/SAANode\\
    association             & Leads to a related Node\\
    yes/no                  & Leads to the binary options following a DecisionNodeYN\\
    additionalInformation   & Leads to additional Information on the current node\\
    \bottomrule
  \end{tabular}
\end{table}

As for relationship types within the graph, there are five distinct ones (Table \ref{table:relationshiptypes}): 
The main relationship the graph operates on, is the priority relationship (Table \ref{table:relationshiptypes}, "R"). It is the most common relationship and portrays the advancement through the graph, and serves to keep certain orders from the modeled manual for rescue operations\cite{treatmentpdf} intact. The R-relationships are numbered, directional relationships, such as ”R1”, ”R2”, ”R3”,... in which R1 is the highest priority. For the lowest priority, ”Rn” is used to signal a relationship path only be taken when no other, higher priority one, is viable. Besides graph logic, these numbers also dictate the order that child nodes will be displayed on the graphic user interface. This ensures certain priorities in treatments to be kept intact, and makes it easier for the treatment personnel to pick the most likely node. Besides the priority relationship, there are the ”yes” and ”no” relationships that specifically only occur after DecisionNodesYN, showing the binary nature of the node. A DecisionYNNode typically does not have any other relationships besides yes and no, since those are the only logical options to be taken based on the question posed in the node. Besides the relationships above used to navigate the main path, there are some that make navigating the entire KG and its paths easier: “BPR”, which leads to related treatment paths and “SAA”, which leads to related standard procedures. Alternatively to leading to specific treatments, BPR and SAA relationships can also be used to quickly access a full list of all BPR and SAA for a manual selection. The “association” relationship meanwhile connects related nodes, such as different ECG nodes across various treatment paths. In that way, treatment personnel can easily jump to a different treatment path depending on the result of the ECG. Besides these nodes for quick navigation, there is one last relationship type that holds the functionality to point out hints and warnings regarding the current treatment: The "additionalInformation” relationship. It can be selected by the user if required, and serves to connect to DisplayNodes holding important information, elaborating for example on contraindications for a treatment, dose calculation for a drug, or things to keep in mind during a standard procedure.

\subsection{Properties} 
\begin{table}
  \caption{Properties}
  \label{table:properties}
  \begin{tabular}{ccl}
    \toprule
    Property Name & Description \\
    \midrule
    Name/id   & Hold names and ids of node for identification\\
    BPR/SAA                     & Hold BPR/SAAs node belongs to\\
    d\_type/value             & Hold information on what needs to be requested from the middleware\\
    min/max                  & Hold information on ranges values should optimally be in\\
    \bottomrule
  \end{tabular}
\end{table}

Every node is uniquely described by properties, and recognizable by an id and the name it has. The “BPR” and “SAA” properties simply declare which path a node belongs to. A change in paths can be realized with these properties, and user can be prompted via the GUI to confirm they want to leave the currently selected path. The properties  d\_type, value and min/max completely serve the communication with the middleware, and allow the KG to make decisions based on the feedback. d\_type values alert the KG to whether a specific value to be requested from the middleware, while “value” declares which value needs to be retrieved, such as “PULSE”. The min and max entries then determine in which range the values should be, and can therefore allow the KG to immediately react to diverting value by either making an automatic decision, or prompting the user. After exploring the exact build of the KG, a short example of how it is used will now be explained before the paper moves on to the point of discussion.

\subsection{Use Case: Inter-module Data Retrieval for smart knowledge management and decision making}

\begin{figure}
\centerline{\includegraphics[width=\textwidth]{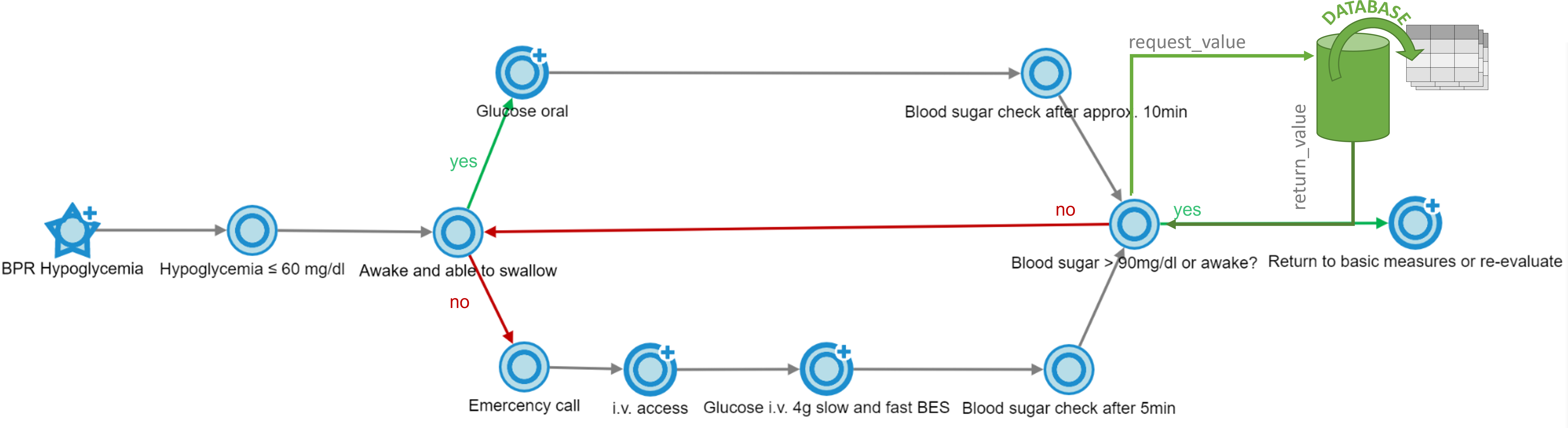}}
\caption{This figure shows a simplified data retrieval process through the middleware and database. It displays a treatment path of the graph, which starts with hypoglycemia being below or equal to 60 mg/dl. This leads first responders to check for awakeness and the ability to swallow for patients. If applicable, the patient will be treated with an oral intake of glucose, followed with a recheck of the patient's blood sugar. The data for this binary decision can then be retrieved from the database, also portrayed in the figure, which directly presents the value to the first responders to enable a faster process.}
\label{usecase2}
\end{figure}

Medical decision-making is a union of the first responders' medical knowledge, the situation at hand, and the patient's vitals. To achieve an inter-communicating system, multiple modules have been developed (Fig. \ref{scheme}) to accomplish medical data transfer. The presented use case (Fig. \ref{usecase2}) shows a simplified view of such a decision being made within the KG. The selected sub-graph shows that the patient in need has a low blood sugar level, which results in a hypoglycemia state. The chosen path is therefore the treatment path for hypoglycemia. To increase the blood sugar level, glucose is needed to be given.
To check whether the patient is doing better or whether another dosage needs to be given, the blood sugar level needs to be evaluated. This checking can be achieved with medical devices first responders have at hand and which record patient vitals. The vitals are then recorded and stored in a KIRETT database, which allows inquiry for the latest data. Newer blood sugar values can then be retrieved on the display \cite{uxNadeem} of the embedded platform and can be manually approved by the medical expert at hand. Such values can then be retrieved in case of DecisionYN-Nodes (section \ref{nodetypes}), where decisions have to be made by experts based on the data they receive. This implementation allows for a fast data evaluation, seeing the target value of the blood sugar level and the at runtime recorded medical blood sugar. In addition to that, this information can also be used to allow the making of accurate decisions as it is based on the manual for rescue operations \cite{treatmentpdf} and supports the report management of emergency services. 

\section{Evaluation}
\begin{table}
  \caption{Evaluation plan for KG-based quality evaluation}
  \label{table:eval}
  \begin{tabular}{p{2.1cm}p{4.9cm}p{6.3cm}}
    \toprule
     Quality criteria &  Description  & Status \\
    \midrule
        Accuracy & Up-to-date behaviour of KG & 37 BPR and 39 SAA included in KG \\
        Completeness & Coverage of all possible nodes and relations & 3046 nodes and 4467 relations covered \\
        Consistency & Follow up of standardized format and scheme & Use of standardized KG format and representation schemes (e.g. node types)\\   
        Continuous Improvement & Monitoring and KG improvement & KG has been discussed and evaluated in regular expert feedback sessions\\
        Explainability & Comprehensive understanding of KG and modules & Step-by-step separation of complex treatments allows comprehensive understanding of KG \\
        Performance & Performance of KG-based queries & Implementation successful via neo4j graph database and Graphlytics visualization\\
        Usability & Navigation on terminal based Application & will be performed in later stage of KIRETT project \\
    \bottomrule
  \end{tabular}
\end{table}

The implementation of the KG is suited and fit for supporting first responders in the treatment recommendation task and representing knowledge in a comprehensive way to the user. Various measures were taken to evaluate the accuracy, completeness, consistency, continuous improvement, explainability (here mainly the understandability of provided information for users), performance and usability of the presented KG. Table \ref{table:eval} describes the criteria and its archieved status for the projects KG. The following sections will present the already satisfied and the future evaluation measures for the graph of this project:

\subsection{Evaluation based on technical and functional requirements}
The accuracy of the KG was fulfilled due to its construction via a manual for rescue operations. In total 37 BPRS and 39 SAA were extracted and implemented and discussed in continuous re-evaluation talks with medical experts. Those approved the correctness of the manual of rescue operations, its construction as a KG (accuracy), and its up-to-date behavior for future improvements (completeness and continuous improvement). Various use-case meetings were used to evaluate certain implementation questions and to settle medical questions to provide optimal accuracy of the graph. The graph was constructed using a standardized format to enhance the consistency of the graph. Complex treatments are separated in step-by-step elements, to enhance the understanding of the treatment and to allow an accurate graph navigation.

\subsection{Future Evaluation of the KG with medical experts}
Performance and usability tests are planned in the upcoming phases of the project. Initially, expert interviews will be conducted to evaluate the effectiveness and performance of the SD and KG. Subsequently, those tools will be fed with random-generated data which will enable a direct test with healthcare workers, to evaluate the usability in simulated real-world scenarios. These tests will be integrated into ongoing education seminars for rescue operators. In addition to that, a quantitative questionnaire will be conducted to prove the usability of the embedded wearable. Future work will present the outcomes of the project in more detail.

\section{Conclusion}
This paper presents a knowledge-graph-based treatment assistant, which provides medical support for rescue operations with the help of treatment recommendations. It provides insights into the KIRETT project and shows how knowledge is retrieved and constructed in a KG. It shows which implementation measures in the graph are needed and how those adjustments interact with each other. Subsequently, it provides examples, to explain the rescue operation and provides evaluation measures for upcoming phases of the project.

\section{Acknowledgements}
 This research received financial support from the Federal Ministry of Education and Research, Germany. Further support and assistance is given, by the KIRETT project coordinator CRS Medical GmbH (Aßlar, Germany) and mbeder GmbH (Siegen, Germany). The authors would like to thank the affiliated partners, of this project, namely Kreis Siegen-Wittgenstein, City of Siegen, the German Red Cross Siegen (DRK), and the Jung-Stilling-Hospital in Siegen.

\bibliographystyle{unsrt}  
\bibliography{references}

\begin{thebibliography}{10}

\bibitem{overviewZenkert}
Johannes Zenkert, Christian Weber, Mubaris Nadeem, Lisa Bender, Madjid Fathi,
  Abu~Shad Ahammed, Aniebiet~Micheal Ezekiel, Roman Obermaisser, and Maximilian
  Bradford.
\newblock Kirett-a wearable device to support rescue operations using
  artificial intelligence to improve first aid.
\newblock In {\em 2022 IEEE International Smart Cities Conference (ISC2)},
  pages 1--4. IEEE, 2022.

\bibitem{abu2022transferrable}
Hasan Abu-Rasheed, Christian Weber, Johannes Zenkert, Mareike Dornh{\"o}fer,
  and Madjid Fathi.
\newblock Transferrable framework based on knowledge graphs for generating
  explainable results in domain-specific, intelligent information retrieval.
\newblock In {\em Informatics}, volume~9, page~6. MDPI, 2022.

\bibitem{frameworkAbuRasheed}
Hasan Abu-Rasheed, Mubaris Nadeem, Mareike Dornhöfer, Johannes Zenkert,
  Christian Weber, and Madjid Fathi.
\newblock Texkg in health domain: The application of knowledge graph based
  framework for explainable recommendations in the contexts of elderly care,
  mental health, and emergency responses.
\newblock In M.R. Alam and Madjid Fathi, editors, {\em Integrated System Design
  and Technology}. Springer, 2023.

\bibitem{uxNadeem}
Mubaris Nadeem, Johannes Zenkert, Christian Weber, Madjid Fathi, and Muhammad
  Hamza.
\newblock Smart ux-design for rescue operations wearable-a knowledge graph
  informed visualization approach for information retrieval in emergency
  situations.
\newblock In {\em 2023 IEEE International Conference on Electro Information
  Technology (eIT)}, pages 180--185. IEEE, 2023.

\bibitem{cardioShad}
Abu~Shad Ahammed, Micheal Ezekiel, and Roman Obermaisser.
\newblock A novel analysis of performance and inference time of machine
  learning models to detect cardiovascular emergency situations of rescue
  patients.
\newblock In {\em 2022 International Conference on Artificial Intelligence of
  Things (ICAIoT)}, pages 1--6, 2022.

\bibitem{respiratoryShad}
Abu~Shad Ahammed, Sampada~Reddy Donthireddy, and Roman Obermaisser.
\newblock Detection of respiratory emergency situation of rescue patients with
  machine learning algorithms.
\newblock In {\em IECON 2022 – 48th Annual Conference of the IEEE Industrial
  Electronics Society}, pages 1--6, 2022.

\bibitem{neuroShad}
Abu~Shad Ahammed, Aniebiet~Micheal Ezekiel, and Roman Obermaisser.
\newblock Time-efficient identification procedure for neurological
  complications of rescue patients in an emergency scenario using
  hardware-accelerated artificial intelligence models.
\newblock {\em Algorithms}, 16(5):258, 2023.

\bibitem{cardioMike}
Aniebiet~Micheal Ezekiel and Roman Obermaisser.
\newblock Time-optimized detection of cardiovascular complications with
  artificial intelligence in rescue operations using fpga-based wearable.
\newblock In {\em 2023 5th International Congress on Human-Computer
  Interaction, Optimization and Robotic Applications (HORA)}, pages 1--8, 2023.

\bibitem{Tommasini2020}
Riccardo Tommasini, Paul Groth, and empty Juan.
\newblock {\em Knowledge Graphs}, pages 1--7.
\newblock Springer International Publishing, Cham, 2020.

\bibitem{abu2023healthcare}
Bilal Abu-Salih, Muhammad Al-Qurishi, Mohammed Alweshah, Mohammad Al-Smadi,
  Reem Alfayez, and Heba Saadeh.
\newblock Healthcare knowledge graph construction: A systematic review of the
  state-of-the-art, open issues, and opportunities.
\newblock {\em Journal of Big Data}, 10(1):81, 2023.

\bibitem{borowiecki2006graph}
Mieczys{\l}aw Borowiecki, John~W Kennedy, and Maciej~M Syslo.
\newblock {\em Graph Theory: Proceedings of a Conference Held in Lagow, Poland,
  February 10-13, 1981}, volume 1018.
\newblock Springer, 2006.

\bibitem{treatmentpdf}
{Arbeitsgruppe Ärztlicher Leiter Rettungsdienst}, {Feuerwehr Siegen}, and {DRK
  Rettungsdienst Siegen-Wittgenstein }.
\newblock {\em Behandlungspfade und Standardarbeitsanweisungen für den
  Rettungsdienst im Kreis Siegen-Wittgenstein}.
\newblock Kreis Siegen-Wittgenstein, Siegen, Germany, 2020.

\end{thebibliography}

\end{document}